\documentclass[lettersize,journal]{IEEEtran}

\usepackage{amsmath,amsfonts}
\usepackage{graphicx}
\usepackage[caption=false,font=normalsize,labelfont=sf,textfont=sf]{subfig}
\usepackage{url}
\usepackage{cite}

\begin{document}

\title{Explore the Ideology of Deep Learning in ENSO Forecasts}

\author{Yanhai Gan, Yipeng Chen, Ning Li, Xingguo Liu, Junyu Dong, and Xianyao Chen
\thanks{This work was supported by the National Science and Technology Major Project of China under Grant 2022ZD0117201, the Natural Science Foundation of China under Grant 42394130 and 42406192, the Natural Science Foundation of Shandong Province under Grant ZR2023ZD38, and the Taishan Scholars Program under Grant TSPD20240807.}
\thanks{Y. Gan, N. Li, X. Liu, and J. Dong are with the State Key Laboratory of Physical Oceanography, Ocean University of China, Qingdao, 266100, China (e-mail: dongjunyu@ouc.edu.cn).}
\thanks{Y. Chen and X. Chen are with the Frontier Science Center for Deep Ocean Multispheres and Earth System, and the Physical Oceanography Laboratory, Ocean University of China, Qingdao, 266100, China (e-mail: chenxy@ouc.edu.cn).}
\thanks{Yanhai Gan and Yipeng Chen contributed equally to this work.}
\thanks{Corresponding authors: Junyu Dong; Xianyao Chen.}
}

\markboth{Journal of \LaTeX\ Class Files,~Vol.~XX, No.~XX, Month~202X}
{Gan \MakeLowercase{\textit{et al.}}: Explore the Ideology of Deep Learning in ENSO Forecasts}

\maketitle

\begin{abstract}
The El Niño–Southern Oscillation (ENSO) exerts profound influence on global climate variability, yet its prediction remains a grand challenge. Recent advances in deep learning have significantly improved forecasting skill, but the opacity of these models hampers scientific trust and operational deployment. Here, we introduce a mathematically grounded interpretability framework based on bounded variation function. By rescuing the “dead” neurons from the saturation zone of the activation function, we enhance the model's expressive capacity. Our analysis reveals that ENSO predictability emerges dominantly from the tropical Pacific, with contributions from the Indian and Atlantic Oceans, consistent with physical understanding. Controlled experiments affirm the robustness of our method and its alignment with established predictors. Notably, we probe the persistent Spring Predictability Barrier (SPB), finding that despite expanded sensitivity during spring, predictive performance declines—likely due to suboptimal variable selection. These results suggest that incorporating additional ocean-atmosphere variables may help transcend SPB limitations and advance long-range ENSO prediction.
\end{abstract}


\section{Introduction}
\IEEEPARstart{T}{he} most significant interannual climate variability signal on Earth—the El Niño-Southern Oscillation (ENSO)—is a typical climate phenomenon resulting from large-scale ocean-atmosphere coupling in the tropical Pacific \cite{cai2018increased, mcphaden2006enso}. The ENSO signal not only affects tropical \cite{bjerknes1969atmospheric} and subtropical \cite{huang2004recent} ocean-atmosphere interactions, but it also leads to frequent global extreme weather events \cite{trenberth1998progress}. Consequently, ENSO has profound impacts on global climate \cite{alexander2002atmospheric, yang2018el}. Therefore, improving ENSO forecasting capabilities holds significant scientific value and socio-economic importance in addressing global climate change.

In recent years, with the continuous development of big data and deep learning in varied fields \cite{vailaya2001image, hirschberg2015advances, esteva2019guide}, some scholars have also started to apply deep learning to ENSO forecasting. Ham et al. \cite{ham2019deep} utilized simulation data and transfer learning to train the convolutional neural network (CNN) \cite{krizhevsky2012imagenet}, and obtained higher prediction performance than the previous mainstream methods. Petersik et al. \cite{petersik2020probabilistic} utilized a small number of predictor variables and samples to train a neural network and obtained similar prediction results.

In spite of these achievements, we are interested in understanding how these deep learning models make decisions to achieve such good prediction performance, which may reveal some known or unknown physics in the ocean, and may also help to further improve the prediction ability. The inherent “black box” nature of AI models makes their decision-making processes difficult to trace, which, despite their exceptional forecasting capabilities, severely limits the credibility of research findings due to the opacity of their decision logic. Therefore, enhancing model interpretability is crucial for building trust in AI-based climate forecasting.

Among various methods, representation visualization is one of the most intuitive post hoc interpretability methods, which can present the features learned by the model and the correlation between inputs and outputs by heatmaps \cite{ancona2017towards, fong2017interpretable, fong2019understanding, kim2019why, wagner2019interpretable}. Through heatmaps, we can intuitively understand the main basis for the model to make the final decisions. There have been many types of representation visualization methods, such as perturbation-based methods \cite{ancona2017towards, fong2017interpretable, fong2019understanding, wagner2019interpretable}, backpropagation-based methods \cite{baehrens2012how, simonyan2013deep, zeiler2014visualizing, springenberg2014striving, bach2015on}, class activation mapping methods \cite{zhou2016learning, selvaraju2017grad, chattopadhay2018grad, wang2020score}.

Perturbation-based method is similar to a causal process, where a certain part of the input is masked, modified, etc., and then fed into the model to get the corresponding output. The importance of this part of the input is measured based on the degree of change in the output \cite{wagner2019interpretable}. Backpropagation-based method maps outputs to inputs layer by layer according to the designed backpropagation rules to get the magnitude of the contribution of the input variables to the outputs \cite{kim2019why}. Class activation mapping methods argued that the final convolutional layer contains the semantic information related to decision making, and by weighting and stacking its feature maps onto the original image, they can visualize the regions that contribute to the target category \cite{selvaraju2017grad}.

Most interpretability methods are developed for image classification models and are not readily applicable to ENSO index regression tasks. These approaches typically assume that features lead to larger output values are the basis for the model's decision-making \cite{chattopadhay2018grad, wang2020score}. While this assumption holds in classification—where a higher output value produces greater class confidence—it breaks down in regression settings. In regression, a larger output does not necessarily imply closer proximity to the expected value. This highlights the inadequacy of such interpretability assumptions for regression models and calls for methods that better reflect their underlying ideology.

We hope for an interpretability method that provides a clear mathematical meaning for the interpretation of ENSO prediction models, thus gaining the trust of researchers and further promoting ENSO research. Essentially, ENSO index forecasting differs fundamentally from conventional computer vision tasks. In this case, it is necessary and hopeful to propose an interpretability approach specifically tailored to the ENSO index forecasting model.

To this end, we first define the principled objective of interpretation: identifying geographic regions most critical for ENSO forecasting. In line with the principled objective, we develop the interpretability method by treating the ENSO forecasting model as a bounded variation function. We quantify the responsibility of each variable for the variance of the function as an importance metric. This approach ensures that the design of the interpretability method is both clear and grounded in mathematics. Ultimately, the proposed method is tested and validated through extensive controlled experiments, and the insights derived from the interpretations are presented to inspire further understanding of ENSO predictability—an outcome we indeed hope to achieve.

\section{Materials and Methods}

\subsection{Data and Convolutional Neural Network}
Ham et al. \cite{ham2019deep} leveraged both climate model simulations and historical reanalysis data for training a deep learning model aimed at multi-year ENSO forecasting. To overcome the limitations imposed by the relatively short observational record, a transfer learning approach was adopted. The CNN was first pre-trained on a large volume of data from the historical simulations of 21 different Coupled Model Intercomparison Project phase 5 (CMIP5) models. Following this pre-training phase, the learned features were transferred, and the model was subsequently fine-tuned using the Simple Ocean Data Assimilation (SODA) reanalysis dataset covering the period 1871-1973. This two-stage training strategy allowed the model to learn robust features from the extensive simulations and then adapt them to the characteristics of the historical reanalysis. The final model performance was evaluated on an independent validation period using the Global Ocean Data Assimilation System (GODAS) reanalysis from 1984 to 2017. This study only used GODAS data for the analysis experiment.

The applied CNN architecture consisted of an input layer receiving Sea Surface Temperature (SST) and upper Ocean Heat Content (HC) anomaly maps for three consecutive months, followed by three convolutional layers interspersed with two max-pooling layers for hierarchical feature extraction and dimensionality reduction, a fully connected layer for integrating learned features, and an output layer predicting the 3-month averaged Nino3.4 index up to 23 months ahead. Four configurations of this CNN, varying in the number of filters and neurons, were trained, and their predictions are averaged for the final result.

\subsection{Total Variation}
Let's simplify the problem first. Suppose we have a univariate function $y = f(x)$. If $f$ is a constant function, then changes in $x$ will not affect the output $y$. In this case, we can consider $x$ as not critical for $f$ at all. If $f$ is not a constant function, a change in $x$ may lead to a change in the output $y$. In this case, the problem is how much responsibility $x$ does have for the changes in $f$'s output $y$. We can measure the magnitude of this responsibility using the total variation of $f$. If the total variation of $f$ is large, the values of $f$ within the domain of $x$ are unpredictable when $x$ is unknown. Conversely, if the total variation of $f$ is small, the values of $f$ within the domain of $x$ change negligibly, and plausible predictions can be made even when $x$ is unknown. Before introducing the proposed method, we review the concept of total variation.

Let $f(x): [a,b] \to \mathbb{R}$ be a real-valued function. If there exists a non-negative constant $K \geq 0$, such that for any partition of $[a,b]$,
\begin{equation}
\Delta: a = x_0 < x_1 < \cdots < x_n = b,
\end{equation}
the following inequality holds:
\begin{equation}
\sum_{i=1}^{n} |f(x_i) - f(x_{i-1})| \leq K,
\end{equation}
then $f(x)$ is said to be a bounded variation function on $[a,b]$, and its corresponding total variation is
\begin{equation}
\text{TV}(f) = \sup_{\Delta} \sum_{i=1}^{n} |f(x_i) - f(x_{i-1})| = \int_{a}^{b} |df(x)|.
\end{equation}
Total variation can be used to measure how fast a univariate function changes and how much the independent variable influences the dependent variable.

\subsection{Practical Partial Total Variation}
Given that $f'(x) = \frac{df(x)}{dx}$, we can convert Eq. (3) into the following form:
\begin{equation}
\text{TV}(f) = \int_{a}^{b} |df(x)| = \int_{a}^{b} |f'(x)dx| = \int_{a}^{b} |f'(x)|dx
\end{equation}
For ENSO prediction, we can consider the regression model as a high-dimensional bounded variation function with $n$ variables: $f(x_1, x_2, \dots, x_n)$. The total variation of this function measures the oscillation of $f$ over all inputting variables, but what we want here is to quantitatively assess the impact of individual variables on the output of $f$. Therefore, we propose a derivative formula of the total variation in Eq. (4). Specifically, we define the partial total variation (PTV) for a single inputting variable $x_1$ as
\begin{equation}
\text{PTV}(f|x_1) = \int \cdots \int_{D} \left| \frac{\partial f}{\partial x_1} \right| dx_1 dx_2 \cdots dx_n.
\end{equation}
In Eq. (5), $\int \left| \frac{\partial f}{\partial x_1} \right| dx_1$ is the total variation of the function $f$ over $x_1$ when all the other inputting variables are fixed (in this case, $f$ can be viewed as a univariate function of $x_1$). To consider the comprehensive effect of $x_1$ along with various configurations of the remaining inputting variables, we integrate over all the inputting variables in Eq. (5).

The defined partial total variation in Eq. (5) quantify the theoretical influence of each inputting variable on the function's output, but it does not account for the uneven distribution of the data samples over the high-dimensional space, which can lead to a large gap between the theoretical results and the actual observations. For example, certain configurations of the inputting variables may never appear in practice and never be seen by the regression model, so the model may have extreme gradients on the data points corresponding to these configurations because they do not contribute to the training and validation phase of the model. In this case, the calculated PTV would be very impractical, because the model does not consider these inputting configurations at all. In addition, one problem is how to distinguish the importance of two inputting variables with the same PTV. In practice, an inputting variable may induce significant changes in the output even when varying within its normal range, while another inputting variable might only trigger substantial output variations when reaching relatively rare extreme values. This implies that the first variable can frequently drive output fluctuations, whereas the second variable influences the output infrequently and conditionally. From a holistic perspective, we consider the first inputting variable more important than the second inputting variable. Consequently, we involve the data distribution $P(x_1, x_2, \dots, x_n)$ when calculating the partial total variation and get the Practical Partial Total Variation (PPTV)
\begin{equation}
\text{PPTV}(f|x_1) = \int \cdots \int_{D} P(x_1, \dots, x_n) \left| \frac{\partial f}{\partial x_1} \right| dx_1 \cdots dx_n
\end{equation}

In Eq. (6), PPTV weights each data point by its probability. In this manner, data points that never appear in practice contribute zero to the result, and frequently occurring data points have more weight than infrequently occurring data points. As a result, PPTV's performance will be more in line with actual observations. However, since we cannot obtain the expression of the probability density and the function f, we cannot analytically solve the integral in Eq. (6) by finding the antiderivative of the integrand. Therefore, we use the Monte Carlo method for approximation of Eq. (6). According to the strong law of large numbers, we can get that
\begin{align} 
\text{PPTV}(f|x_1) &= \int \cdots \int_{D} P(x_1, \dots, x_n) \left| \frac{\partial f}{\partial x_1} \right| dx_1 \cdots dx_n \nonumber \\
&= \mathbb{E}_{P(x_1, \dots, x_n)} \left[ \left| \frac{\partial f}{\partial x_1} \right| \right] \nonumber \\
&\approx \frac{1}{m} \sum_{i=1}^{m} \left| \frac{\partial f(x^i)}{\partial x_1^i} \right|. \label{eq:pptv_approx}
\end{align}
In Eq. (\ref{eq:pptv_approx}), $m$ denotes the number of samples, and $\left| \frac{\partial f(x^i)}{\partial x_1^i} \right|$ denotes the value of $\left| \frac{\partial f}{\partial x_1} \right|$ calculated at the data point $x^i := <x_1^i, x_2^i, \dots, x_n^i>$. Because $\left| \frac{\partial f(x^i)}{\partial x_1^i} \right|$ can be calculated by the automatic differentiation, we can approximate the proposed PPTV according to Eq. (\ref{eq:pptv_approx}) efficiently. Particularly, for a certain inputting variable $x_1$, we go through all the training samples to simulate sampling from the distribution of the data, and average all the $\left| \frac{\partial f}{\partial x_1} \right|$ values calculated at the samples to get the PPTV of this variable.

\section{Results}
\subsection{Application and Validation of the Proposed Interpretability Method}
\begin{figure*}[!t]
    \centering
    \includegraphics[width=0.9\textwidth]{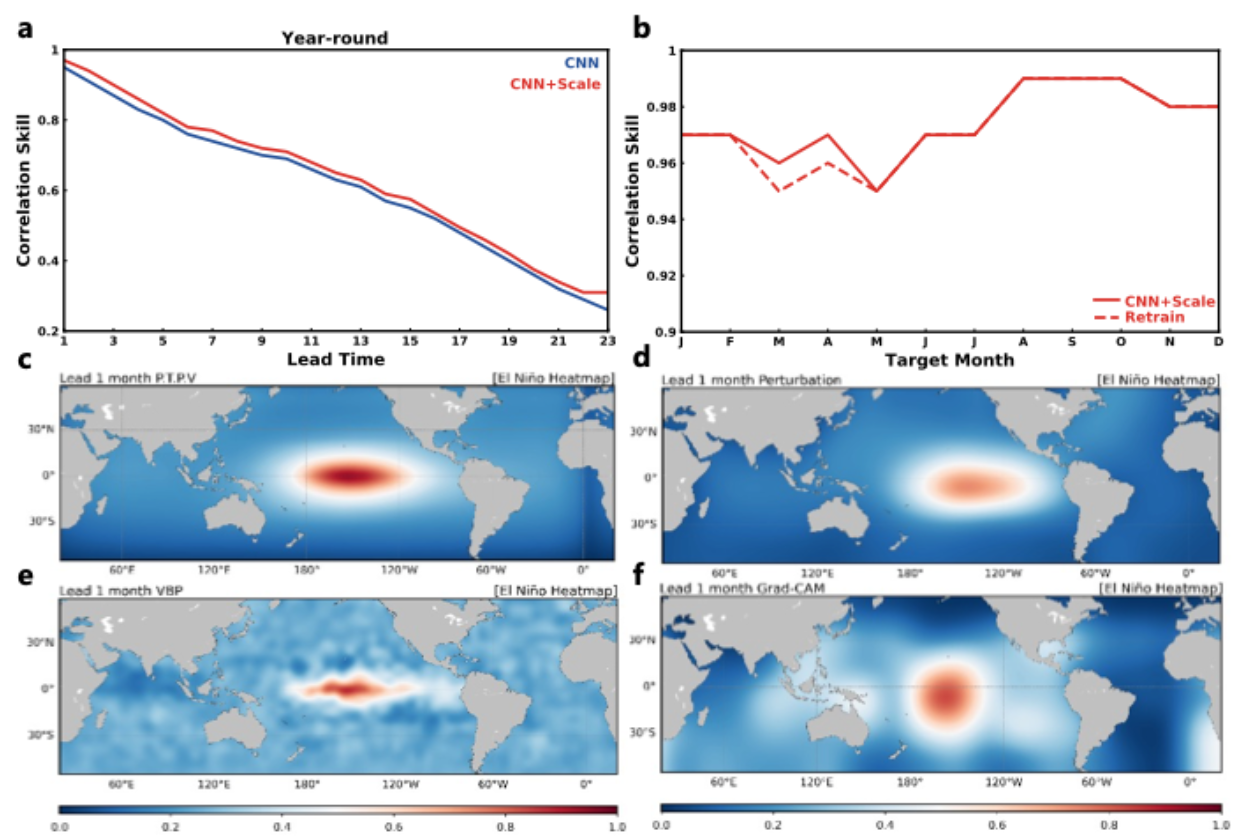}
    \caption{Correlation skill of ENSO forecasting and explanation results of different methods. (a) Correlation skill of ENSO forecasting in Year-round. The blue line is the result of Ham et al. (2019), and the red line is the result of the revised model. (b) The 1 lead time results of the retrained model based on PPTV. The solid red line is the result of the revised model, and the dash red line is the result of the retrained model based on PPTV. (c)-(f) Explanation results of PPTV, Perturbation, VBP, and Grad-CAM, respectively. Values are normalized to the range [0, 1] with 1 indicating higher importance of the region. Note that, for the explanation results at a lead time of 1 month, the 12 target months are averaged to produce a holistic result.}
    \label{fig:fig1}
\end{figure*}

We applied the proposed interpretability method to the forecasting models introduced by Ham et al. \cite{ham2019deep}. However, the first attempt did not provide us with any meaningful results. Through anatomical inspection, we noticed that Ham et al.'s model was seriously plagued by gradient vanishing. Particularly, a large number of hidden features of the model fall into the saturation zone of the employed activation function. This makes the model difficult to analyze using the backpropagation paradigm. To address the issue, we insert learnable parameters before the activation function to adaptively calibrate the features in both spatial and channel dimensions to a manageable range. This remedy effectively resolved the gradient vanishing issue and enhanced the prediction performance (Fig.~\ref{fig:fig1}a).

Recalling that our interpretability objective is defined as identifying the geographic regions most critical for ENSO forecasting, we first need to answer what is meant by ``critical''. Clearly, critical regions should bear the primary responsibility for the variations in ENSO index predictions, in contrast to non-critical regions. To establish a quantitative metric for characterizing this responsibility and to enable its computation, we propose the Practical Partial Total Variation (PPTV, see Methods) method, which includes both the theoretical formulation of PPTV and an approximate calculation approach. As the PPTV method is derived from bounded variation functions and statistical theory, it is mathematically more interpretable than previous heuristic-based methods. Especially, by treating the prediction model as a continuous function, PPTV can be well applied to regression models with no model modification, enabling more fast and accurate identification of key predictors for ENSO forecasting.

We applied the proposed PPTV method to dissect the revised ENSO prediction model, and compared the explanation with the results of other interpretability methods. Specifically, we incorporate the Perturbation \cite{fong2017interpretable}, VBP \cite{simonyan2013deep}, and Grad-CAM \cite{selvaraju2017grad} methods for comparison. Fig.~\ref{fig:fig1}c-f display the results of several interpretability methods with 1 month lead time forecasting which are averaged for the all 12 target months. Obviously, our PPTV method manifests the most concentrated distribution of the important areas. Although Perturbation \cite{fong2017interpretable} obtains similar results as PPTV, it is computationally intensive. The VBP \cite{simonyan2013deep} method produces noisy results that become more severe with increasing lead time. Lastly, the important areas identified by Grad-CAM are more dispersive than the results of PPTV.

To validate the effectiveness of the PPTV method, we employ a retraining methodology. Specifically, we retrain the model using only the data of the most important areas identified by PPTV. The rationale lies in the idea that if we can achieve the same performance as the model trained on the full dataset, we can affirm the effectiveness of the identified predictors. For 1 lead time, we use only the data of areas with PPTV above 0.5, i.e., tropical Pacific Ocean in Fig.~\ref{fig:fig1}c, to retrain the model. The correlation skill of the retrained model is presented in Fig.~\ref{fig:fig1}b. The results indicate that using only the data from the identified predictor areas results in negligible loss of correlation skill, empirically validating the PPTV method. Additionally, this also illustrates that for one-month lead time, the information from the tropical Pacific Ocean is sufficient for the model to make the prediction decision.

\subsection{Attention Extends First Latitudinal and Then Longitudinally}
\begin{figure*}[!t]
    \centering
    \includegraphics[width=0.9\textwidth]{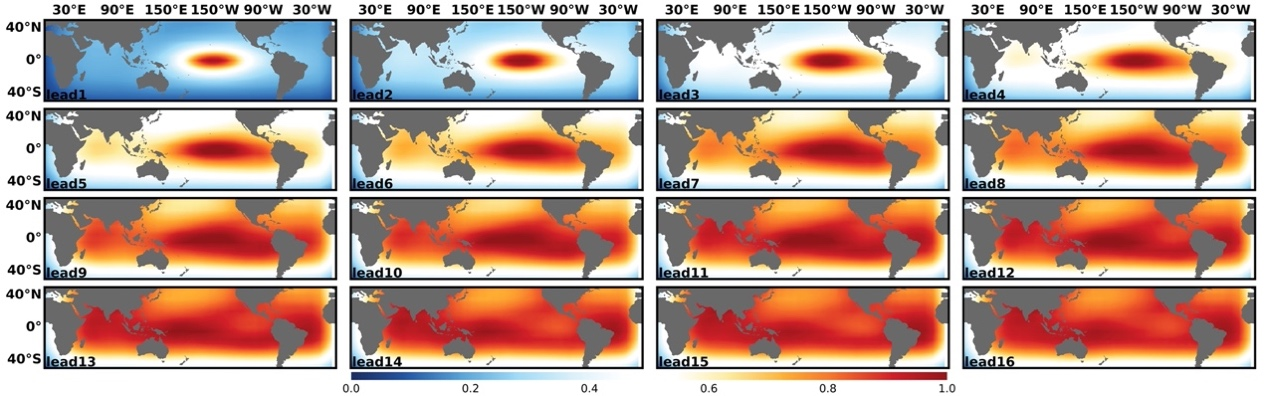}
    \caption{PPTV explanation of the models with the lead times ranging from 1 to 16 months. Values are normalized to [0, 1] and 1 indicates higher importance of the region. Note that the 12 target months are averaged for visual analysis.}
    \label{fig:fig2}
\end{figure*}

To explore how the model's focus changes with different lead times, we apply the PPTV method to visualize the models at various lead times. In the analysis, the results for all 12 target months are averaged to isolate and examine the relationship between PPTV and lead times. Given the poor performance at longer lead times, we limit our analysis to lead times ranging from 1 to 16 months (Fig.~\ref{fig:fig2}). Initially, the model focuses on the tropical Pacific Ocean. As lead time increases, its attention expands to other regions along the same latitude, including the tropical Indian and Atlantic Oceans. At even longer lead times, the model's focus becomes increasingly dispersed. Although not depicted, experiments with longer lead times exhibit similarly scattered attention patterns as observed at the 16-month lead time. Throughout these lead times, areas south of 40$^{\circ}$S consistently receive the least attention.

\begin{figure*}[!t]
    \centering
    \includegraphics[width=0.9\textwidth]{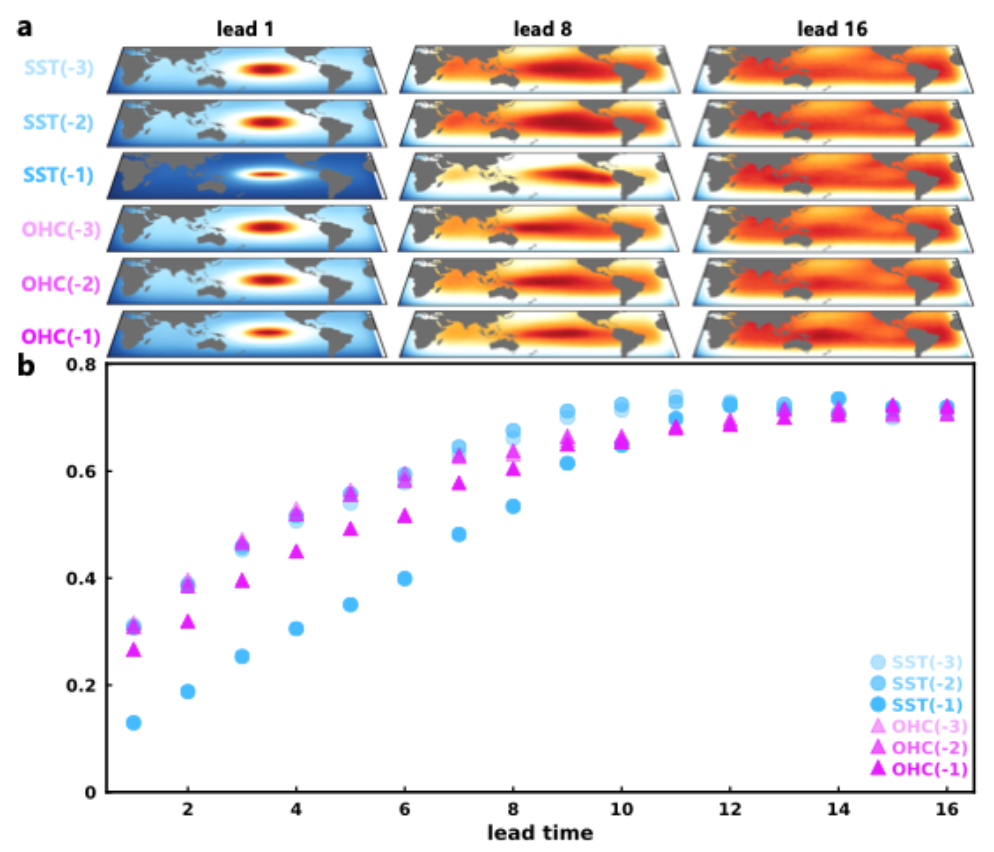}
    \caption{PPTV explanation of the model for each individual channel with the lead times ranging from 1 to 16 months. (a) PPTV visualization of each individual channel, include SST (-1 to -3) and OHC (-1 to -3). (b) PPTV attention of each individual channel of model with the increased lead time.}
    \label{fig:fig3}
\end{figure*}

The PPTV method facilitates multi-scale mechanism analysis of the model through dual operational modes: (1) cross-channel aggregation for coarse-grained spatial mechanism exploration, and (2) channel-specific examination for fine-grained spatiotemporal predictor identification. The forecasting model accepts six channels as input - comprising three consecutive monthly records of SST and ocean heat content (OHC) measurements. Spatial attention distributions derived from channel-averaged weights are visualized in Fig.~\ref{fig:fig1} and Fig.~\ref{fig:fig2}, revealing the model's geographical focus areas in the prediction tasks. However, further analysis of the significance values for different lead times and channels reveals a consistent pattern of increased dispersion as lead time grows (Fig.~\ref{fig:fig3}). At lead time 1, the model displays distinct significance value distributions for SST and OHC over the three consecutive months (Fig.~\ref{fig:fig3}a). At this point, the model focuses more on the variable closest to the target month (-1), with SST (-1) showing a more concentrated attention compared to OHC (-1). When the lead time extends to 8 months, the model exhibits varying attention across different variables. For SST, the model continues to prioritize data from the closest month (-1), but for OHC, the focus that only on data from (-1) weakens, and overall, the model's attention to OHC becomes more concentrated than to SST. As the lead time further increases to 16 months, the model's attention to OHC across the three consecutive months begins to align, followed by a similar alignment for SST. This indicates that the dispersion of attention occurs not only within the three consecutive months of each variable at shorter lead times but also between the two variables at both shorter and longer lead times (Fig.~\ref{fig:fig3}b). The evolution of attention aligns with our physical understanding of ENSO forecasting: SST is more important at shorter lead times, with information closer to the target month being crucial, while OHC becomes more significant at longer lead times. Note that, for clarity in analysis, we used the mean of the significance values (ranging from 0 to 1) across the entire region as the final attention measure (attention indicator, 0-1 scale, Fig.~\ref{fig:fig3}b).

\subsection{Spring Forecasts Need to Capture More Information}
\begin{figure*}[!t]
    \centering
    \includegraphics[width=0.9\textwidth]{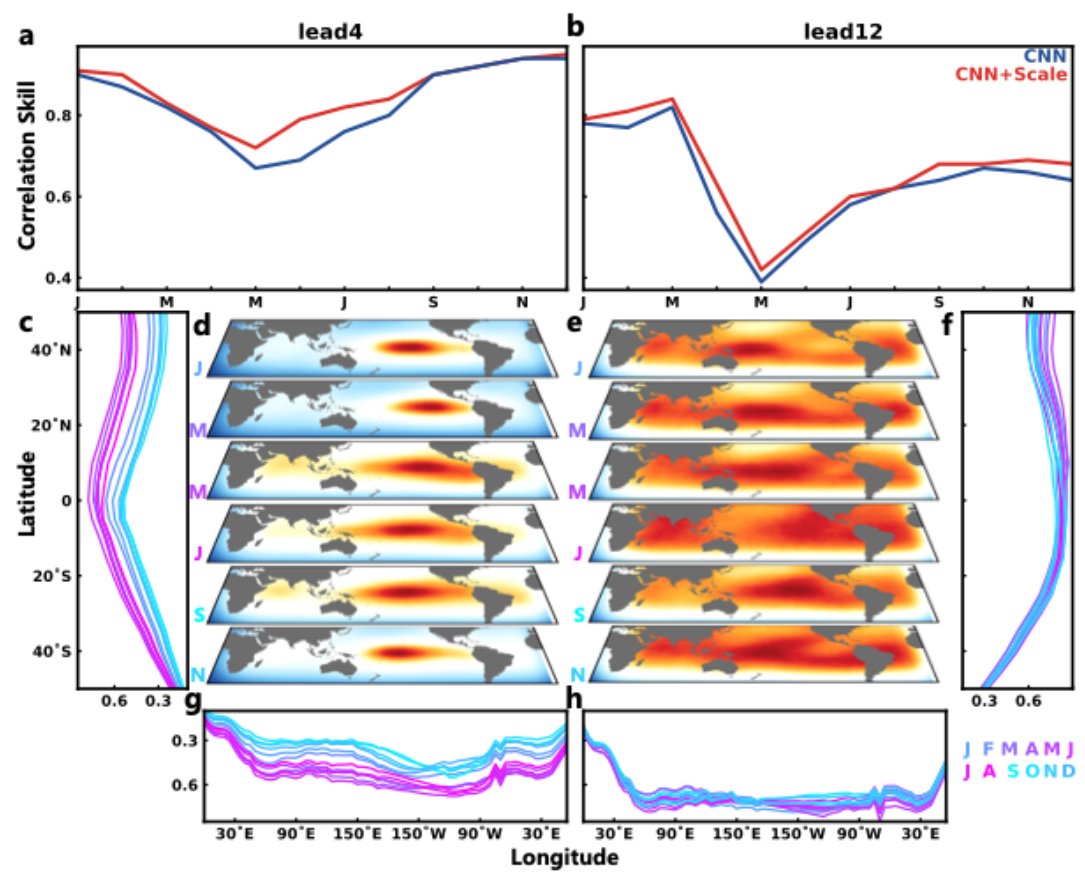}
    \caption{Seasonal PPTV evolution. (a) Correlation skill of ENSO at lead time of 4 months. (b) Same as (a) but for the lead time of 12 months. (d) Seasonal PPTV visualization of lead time of 4 months. (c, g) Seasonal zonal mean and meridional mean of PPTV of lead time of 4 months. (e, f, h) Same as (d, c, g) but for the lead time of 12 months.}
    \label{fig:fig4}
\end{figure*}

The deep learning model has extended the prediction time for ENSO to over a year, partially crossing the Spring Predictability Barrier (SPB). However, predictions for spring months still suffer from reduced performance compared to non-spring months due to the influence of the SPB (Fig.~\ref{fig:fig4}a, b). Using the PPTV method, we analyze how the model's attention changes between spring and non-spring months at different lead times (Fig.~\ref{fig:fig4}c-h, purple and cyan markers). At lead time 4 for spring months, the model's focus expands from just the tropical central Pacific to other regions at the same latitude, including the tropical Indian and Atlantic Oceans (Fig.~\ref{fig:fig4}c, d, g). And for non-spring months, the model's focus shifts from covering the entire tropical region, including all three oceans, back to concentrating primarily on the tropical Pacific. The focus exhibits seasonal regularity evolution. By calculating the mean of the significance values in the zonal (Fig.~\ref{fig:fig4}c) and meridional (Fig.~\ref{fig:fig4}g) directions, we observe that the model's attention is more dispersed during spring months compared to non-spring months (with significant differences in both latitude and longitude). At lead time 12, the meridional pattern (Fig.~\ref{fig:fig4}h) shows slightly more seasonal regularity than the zonal pattern (Fig.~\ref{fig:fig4}f), but not as distinctly cyclic as lead time 4 (Fig.~\ref{fig:fig4}e vs. Fig.~\ref{fig:fig4}d).

\begin{figure*}[!t]
    \centering
    \includegraphics[width=0.9\textwidth]{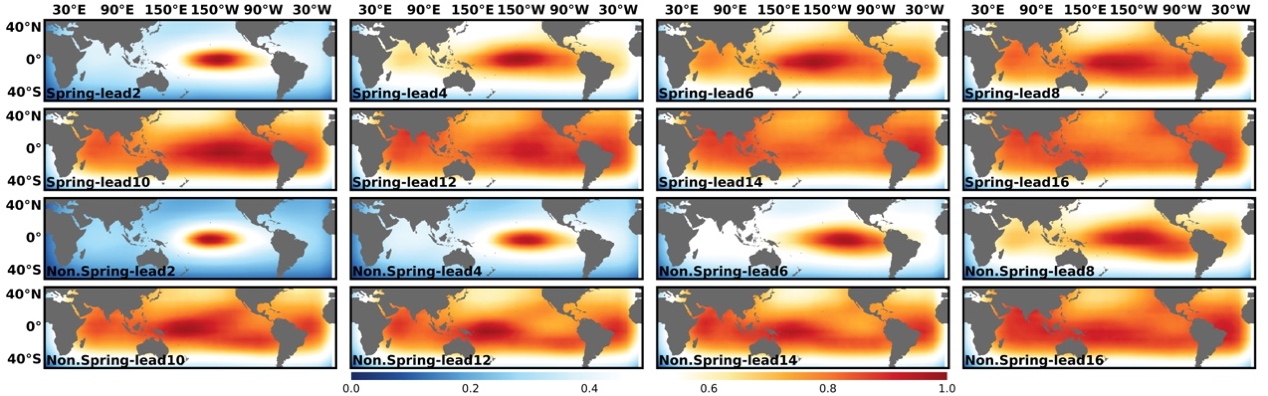}
    \caption{PPTV for spring and non-spring months as lead times increase. Note that the average of March to June represents the spring months, while September to December represents the non-spring months.}
    \label{fig:fig5}
\end{figure*}

We further group the target months into spring and non-spring categories to investigate how attention evolves with increasing lead time (Fig.~\ref{fig:fig5}). At shorter lead times, the attention for spring months becomes more dispersed compared to non-spring months as the lead time increases. At longer lead times, although attention is generally more scattered, non-spring months still exhibit more concentrated areas (high value) of attention than spring months. Moreover, as we previously concluded, the model's attention tends to spread from the tropics to higher latitudes as lead time increases. When analyzing attention evolution at a fixed lead time, we observe a cyclic pattern focused on tropical regions (with spring months expanding attention from the tropical Pacific to the Indian and Atlantic Oceans, while non-spring months focus more on the tropical Pacific). This suggests that, aside from the natural outward spread of attention at longer lead times, the model can effectively address the SPB by concentrating only on tropical regions.

This supports our earlier analysis that the model's attention is always more dispersed during spring months, regardless of lead time. Meanwhile, from the perspective of the training data, one possible explanation is that there are either no or very few samples in the data that provide clear long-term predictability for ENSO. In other words, dynamical models, which are based on existing physical mechanism and parameterization schemes, tend to show more consistent performance when simulating physical processes affecting short-term ENSO prediction. However, they diverge in their ability to simulate long-term physical processes. This highlights an area of future research: using methods like PPTV to explore unknown physical processes that significantly influence long-term ENSO predictability.

\section{Conclusion}
In this paper, we propose an interpretability method PPTV that is suitable for the Ni{\~n}o3.4 regression models. PPTV does not require modifying the model or manually tuning the parameters. Furthermore, the proposed PPTV method is mathematically formulated, making the analysis understandable for oceanologists.

Extensive ablation and controlled experiments demonstrate the effectiveness of PPTV: retrained models using only the discovered significant regions perform nearly as well as those trained on full data, confirming the substantive importance of the identified areas. These regions are primarily located in the tropical Pacific Ocean, with additional influence from the tropical Indian and Atlantic Oceans, especially as lead time increases.

In addition, we dissect the model from multiple perspectives and find that the model focuses on a wider range of areas when the target month is within the spring forecast barrier months. However, even the model grasps more clues from the ambient areas, the prediction performance still decreases significantly. This suggests that the SPB might be an intrinsic component of ENSO dynamical processes, or that the information embedded in the input data is no longer sufficient for the model to make accurate predictions. In the future, more modalities of the ocean and the related atmosphere may be integrated to break through the spring prediction barrier.

\section*{Acknowledgment}
This work was supported by the National Science and Technology Major Project of China under Grant 2022ZD0117201, the Natural Science Foundation of China under Grant 42394130 and 42406192, the Natural Science Foundation of Shandong Province under Grant ZR2023ZD38, and the Taishan Scholars Program under Grant TSPD20240807.

\section*{Data Availability Statement}
The Global Ocean Data Assimilation System (GODAS) reanalysis data were retrieved from the Physical Sciences Laboratory (PSL) at NOAA (https://www.psl.noaa.gov/data/gridded/data.godas.html). The Coupled Model Intercomparison Project phase 5 (CMIP5) dataset is available at the Earth System Grid Federation (ESGF) node (https://esgf-node.llnl.gov/projects/cmip5/). The Simple Ocean Data Assimilation (SODA) version 2.2.4 dataset can be accessed from the NCAR Climate Data Guide (https://climatedataguide.ucar.edu/climate-data/soda-simple-ocean-data-assimilation). The PPTV method proposed and applied in this study has been made publicly available on GitHub at https://github.com/liuxingguo9349/pptv-enso-env.

\bibliographystyle{IEEEtran}
\bibliography{references}

\end{document}